# Learning and evolution: factors influencing an effective combination

**Paolo Pagliuca**

Institute of Cognitive Sciences and Technologies

National Research Council (CNR)

Via S. Martino della Battaglia, 44

00185 Roma, Italia

## Abstract

The mutual relationship between evolution and learning is a controversial argument among the artificial intelligence and neuro-evolution communities. After more than three decades, there is still no common agreement on the matter. In this paper the author investigates whether combining learning and evolution permits to find better solutions than those discovered by evolution alone. More specifically, the author presents a series of empirical studies that highlight some specific conditions determining the success of such a combination, like the introduction of noise during the learning and selection processes. Results are obtained in two qualitatively different domains, where agent/environment interactions are minimal or absent.

**Keywords:** evolution, learning, stochastic hill-climbing, evolutionary strategies

## 1. Introduction

The interplay between learning and evolution has been studied for decades, but it is still a very controversial topic. Despite the huge amount of work, to what extent the interaction between learning and evolution actually fosters the development of successful behaviors is still a matter of debate in the scientific community. Indeed, as it is well described in [1-2], there exist some controversial arguments about the effect of learning on evolution. Some studies revealed how learning accelerates evolution [3-15], while other works demonstrated that learning does not provide any advantage on the course of evolution [16-24].

As explained in [25], «Evolution and learning (or phylogenetic and ontogenetic adaptation) are two forms of biological adaptation that differ in space and time. Evolution is a process of selective reproduction and substitution based on the existence of a population of individuals displaying variability at the genetic level. Learning, instead, is a set of modifications taking place within each single individual during its own lifetime. Evolution and learning operate on different time scales. Evolution is a form of adaptation capable of capturing relatively slow environmental changes that might encompass several generations (e.g., the perceptual characteristics of food sources for a given species). Learning, instead, allows an individual to adapt to environmental modifications that are unpredictable at the generational level. Learning might include a variety of mechanisms that produce adaptive changes in an individual during its lifetime, such as physical development, neural maturation, variation of the connectivity between neurons, and synaptic plasticity. Finally, whereas evolution operates on the genotype, learning affects only the phenotype and phenotypic modifications cannot directly modify the genotype».

Therefore, learning and evolution provide two alternative frameworks [26] to understand the adaptive changes allowing evolving agents to behave more effectively based on the particular environment they are situated in. However, learning and evolution might concur to the development of complex behaviors. The first work trying to highlight the positive effect of learning on evolution is proposed in [6]. The authors considered a simple experimental setting, where genotype and phenotype representations are trivial and their relationship is immediate. The genotype is a string of bits, while the phenotype is a neural network. Given a genotype, a 1-bit corresponds to the presence of a particular connection in the network. Conversely, a 0-bit means that the corresponding connection is absent. In the abstract task used by the authors, only a specific combination of genes (i.e., a genotype with all 1-bits) gets a fitness score of 1, whereas all other genotypes receive a fitness score of 0. To study the effect of the combination of evolution and learning, the authors considered a control situation in which the alleles could also assume a "*" value and learning operated by simply assigning random values to these alleles. The results collected demonstrated how the combination of learning and evolution does permit finding the solution of the problem, while the use of learning or evolution alone fails. To date, however, this method has not been successfully applied to realistic problems, as for example the evolution of robots selected based on their capability to solve a problem that cannot be solved by using evolution alone.

The seminal work in [6] has been a source of inspiration for many other works. Some studies stated that the combination of evolution and learning is advantageous over the application of the single approaches [3-5,8,13-15,27-32], while others showed that learning actually decelerates evolution [16-24,33]. Moreover, some works focused on the analysis of the benefits and limitations of plasticity/learning or the conditions under which learning accelerates/decelerates evolution [7,34-39].

In spite of the huge amount of publications arguing that learning can accelerate or decelerate evolution, there are aspects that have not yet been considered and require a deeper analysis. In particular, two weak points can be found in most of the cited studies. First, the majority of these works do not consider computational costs, a crucial factor in determining the superiority/inferiority of a method over another. Evaluating different algorithms under the same evolutionary conditions and for the same duration is pivotal to shed the light on and untangle this topic. Second, the majority of the conducted studies have been carried out by using trivial and often abstract problems specifically designed to test the algorithms. Instead, the main interest of this work is in verifying whether the combination of learning and evolution might be advantageous over traditional approaches in realistic and more challenging domains.

The learning process used in the presented experiments is obtained through a stochastic hill-climbing process [40] exploring the search space and looking for adaptive solutions. Variations concurring to such adaptive traits are inherited in the population, resulting in Lamarckian learning [41-42]. Lamarckian learning has been successfully used for evolving solutions in multi-agent environments [42], training recurrent neural networks [43-45], training convolutional neural networks for image classification [46], solving pattern classification and function optimization problems [47] and evolving robot's body and brain [48-49]. For the sake of clarity, the author wants to stress that this approach comes from the original work in [6], though its implementation is quite different.

In this paper the author investigates whether the combination of learning and evolution permits to find better solutions than evolution alone. This will be tested in two completely different domains: (i) the well-known 5 bit-parity task and (ii) the popular double-pole balancing problem [50], a benchmark task largely used to compare different evolutionary algorithms [51-54]. Results collected indicate that the combination of evolution and learning leads to the discovery of better solutions than evolution alone, especially with the addition of noise to the learning process. Furthermore, when learning and evolution are combined, the computational cost required to find out a suitable solution is lower.

In the next section the evolutionary tasks used in this work are illustrated. Section 3 contains a description of the different evolutionary algorithms used. In section 4 the results of the performed analysis are presented. Finally, in section 5 the author draws out his conclusions.

## 2. Tasks

This section contains a description of the evolutionary tasks used. It is worth noting that the two considered tasks represent well-known benchmark problems in evolutionary computation. As stated in the introduction, the main reason is to validate the beneficial effect of combining learning and evolution in widely recognized and challenging domains.

### 2.1. 5-bit parity

Digital circuits (Figure 1) are systems computing digital logic functions, like the sum and/or the multiplication of digital numbers. They receive two or more binary (Boolean) values as inputs and produce one or more binary values as output. Digital circuits are made of several logic gates receiving two binary values (from the input pattern and/or from the output of other logic gates) in input and produce one binary value in output. The output of each logic gate is the result of an elementary logic function (AND, OR, NAND, NOR, etc.) of the input. The logic function computed by a circuit depends on the functions computed by its constituent logic gates and how they are wired [55].

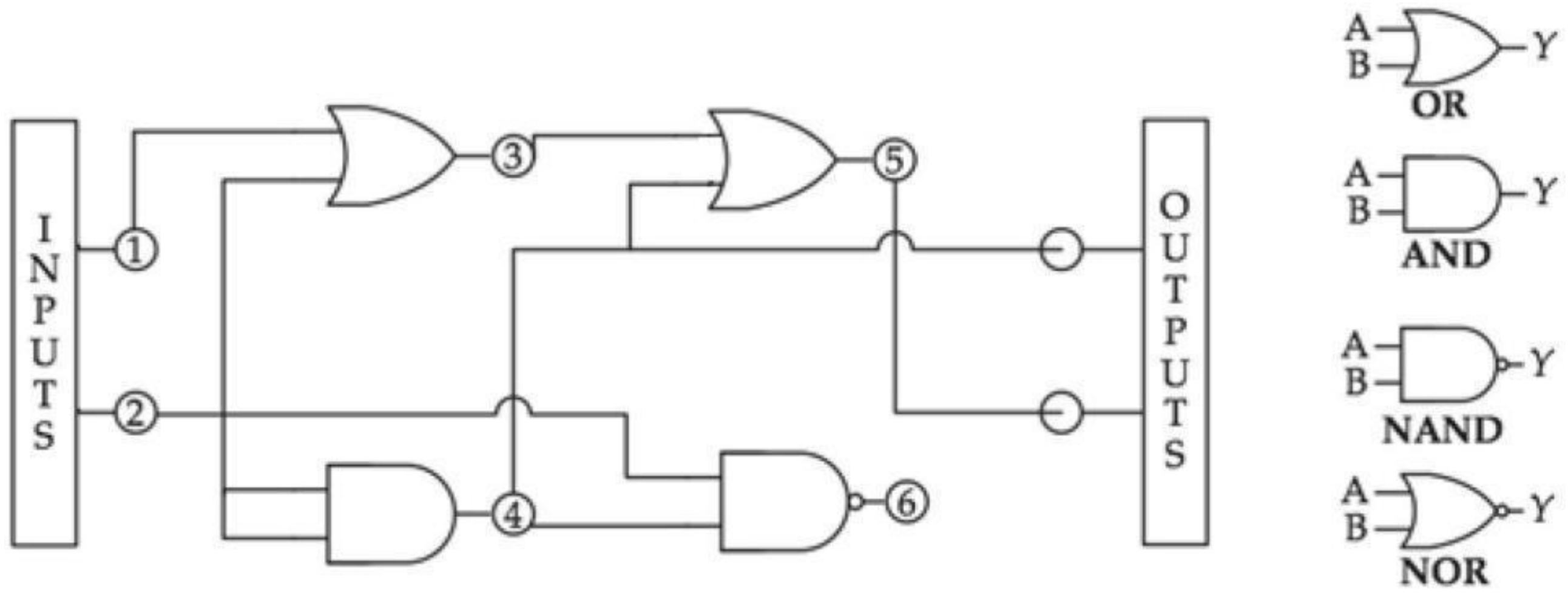

Figure 1. An example of a digital circuit with two inputs, two outputs, and four gates. On the right side there are four symbols corresponding to the four types of usable logic gates. The numbers 1–2 indicate the binary states provided as inputs to the circuit (input pattern). The numbers 3–6 identify the outputs computed by the four corresponding logic gates. The output of the circuit corresponds to the outputs of the two logic gates wired to the output units (in this example the output is given by gates 4 and 5). The lines indicate how gates are wired. Adapted from [55].

Digital circuits can be made in hardware or simulated in a computer. In standard electronic digital circuits, the number and type of gates and how they are wired is hardwired and hand-designed. In reconfigurable electronic digital circuits (such as the FPGA, see [56]), instead, the logic function computed by each gate and how gates are wired can vary. In evolvable hardware applications or, more generally, in evolutionary circuits, the logic function computed by each gate and the way gates are wired are encoded in artificial genotypes and evolved. Evolving circuits are selected depending on their fitness, which is usually computed by measuring to what extent the function computed by a circuit approximates a given target function [57].

In the experiments reported in this paper, the task consists in evolving simulated digital circuits with five inputs, 400 logic gates divided into 20 layers of 20 gates, and one output for the ability to compute a 5-bit even parity function (i.e. to produce as output 1 when there is an even number of 1s in the input pattern and 0 otherwise). This function constitutes a rather difficult problem for evolving circuits including OR, AND, NAND, and NOR logic gates [55,58-59].

Circuits are evaluated for the ability to map the $2^n$ - 32 in this case with n = 5 - possible input patterns into the corresponding desired outputs (i.e., 1 for input patterns with an even number of 1s and 0 otherwise). The fitness of the circuits is calculated on the basis of the following equation:

$$F = 1 - \frac{1}{2^n} \sum_{j=1}^{2^n} \left| O_j - E_j \right| \quad (1)$$

where n is the number of inputs of the circuit, j is the number of the input patterns varying in the range $[1, 2^n]$, Oj is the output of the circuit for pattern j, Ej is the desired output for pattern j.

A circuit evaluation requires one evolutionary step. The evolutionary process is continued until the total number of performed steps exceeds $10^8$, i.e. when around $\frac{10^8}{32} = 3.125 * 10^6$ circuits have been evaluated.

### 2.2. Double-pole balancing

The double-pole balancing problem, introduced in [50], consists in controlling a mobile cart with two poles attached through passive hinge joints on the top of the cart for the ability to keep both poles balanced (see Figure 2). This problem became a commonly recognized benchmark for the following reasons: (i) it involves fundamental aspects of agent's control (e.g., situatedness, non-linearity, temporal credit assignment [60]), (ii) it is intuitive and easy to understand, and (iii) it requires a low computational cost.

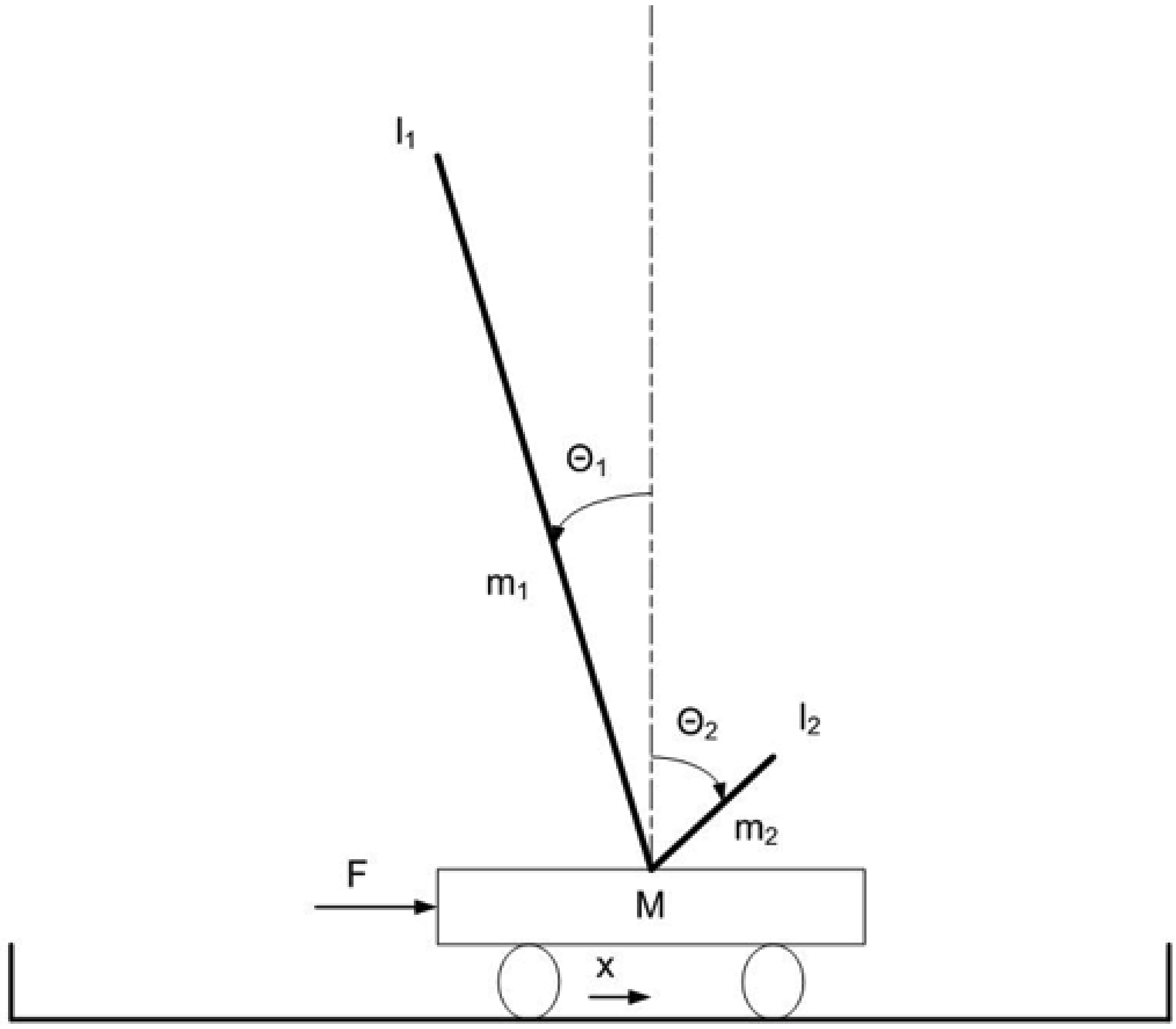

Figure 2. The double-pole balancing problem

The cart has a mass of 1Kg. The long pole has a mass of 0.5Kg and a length of 1.0 m, while the short pole has a mass of 0.05 Kg and a length of 0.1 m. The cart can move along one dimension within a track of 4.8 m. In this paper only the non-Markovian version of the problem is considered, in which the controller is provided with four sensors encoding the current position of the cart on the track ($x$), the current angle of the pole(s) ($\theta_1$ and $\theta_2$) and a constant bias of 0.5. The motor controls the force applied to the cart along the x axis. The goal is to control the force applied to the cart so as to maintain the angle of the poles and the position of the cart within a viable range (details are provided below).

The following equations [50] are used to compute: the effective mass of the poles (2), the acceleration of the poles (3), the acceleration of the cart (4), the effective force on each pole (5), the next angle of poles (6), the velocity of the poles (7), the position of the cart (8), and the velocity of the cart (9), respectively:

$$m\hat{}_i = m_i\left(1 - \frac{3}{4}cos^2\theta_i\right) \quad (2)$$

$$\theta'_i = \frac{-3}{4l_i} \quad (3)$$

$$x' = \frac{F + \sum_{i=0}^{N} F\hat{}_i}{M + \sum_{i=0}^{N} m\hat{}_i} \quad (4)$$

$$F\hat{}_i = m_i l_i \theta'^2_i sin\theta_i + \frac{3}{4} m_i cos\theta_i \left(\frac{\mu_{pi}\theta'_i}{m_i l_i} + g sin\theta'_i\right) \quad (5)$$

$$x[t + 1] = x[t] + \tau x'[t] \quad (6)$$

$$x'[t + 1] = x'[t] + \tau x'[t] \quad (7)$$

$$\theta_i[t + 1] = \theta_i[t] + \tau\theta'_i[t] \quad (8)$$

$$\theta'_i[t + 1] = \theta'_i[t] + \tau\theta'_i[t] \quad (9)$$

where $x$ is the position of the cart on the track that varies in the range [-2.4, 2.4] m, $x'$ is the velocity of the cart, $\theta_i$ is the angular position of the i-th pole, $\theta'_i$ is the angular velocity of the i-th pole. The dynamics of the system was simulated by using the Runge-Kutta fourth-order method. The step size was set to 0.01s.

The controller of the agent is constituted by a neural network with four sensory neurons. The sensory neurons encode the position of the cart ($x$), the angular position of the pole(s) ($\theta_1$ and $\theta_2$). The controller is also provided with a bias input of 0.5. The state of the $x$,$\theta_1$and $\theta_2$ sensors are normalized in the [-0.5, 0.5], [-$\frac{5}{13}$ * $\pi$, $\frac{5}{13}$ * $\pi$] and [-$\frac{5}{13}$ * $\pi$, $\frac{5}{13}$ * $\pi$] ranges, respectively. The activation state of the motor neuron is normalized in the range [-10.0, 10.0] N and is used to set the force applied to the cart. The state of the sensors, the activation of the neural network, the force applied to the cart, and the position and velocity of the cart and of the poles are updated every 0.02 s.

Analogously to [53,61], to promote the evolution of solutions that are robust with respect to the initial position and velocity of the cart and of the poles, each controller is evaluated for 8 trials that varied with respect to the initial state of the system. In a first experimental condition (Fixed Initial States condition) the initial states reported in Table 1 were used. In a second experimental condition (Randomly Varying Initial States condition) the initial states were set randomly during each trial in the range described in Table 2.

| Episode | $x$ | $x'$ | $\theta_1$ | $\theta_2$ | $\theta'_1$ | $\theta'_2$ |
|---|---|---|---|---|---|---|
| 1 | -1.944 | 0 | 0 | 0 | 0 | 0 |
| 2 | 1.944 | 0 | 0 | 0 | 0 | 0 |
| 3 | 0 | -1.215 | 0 | 0 | 0 | 0 |
| 4 | 0 | 1.215 | 0 | 0 | 0 | 0 |
| 5 | 0 | 0 | -0.10472 | 0 | 0 | 0 |
| 6 | 0 | 0 | 0.10472 | 0 | 0 | 0 |
| 7 | 0 | 0 | 0 | -0.135088 | 0 | 0 |
| 8 | 0 | 0 | 0 | 0.135088 | 0 | 0 |

Table 1. Initial states used during different trials carried out in the Fixed Initial States condition.

| | min | max |
|---|---|---|
| $x$ | -1.944 | 1.944 |
| $x'$ | -1.215 | 1.215 |
| $\theta_1$ | -0.10472 | 0.10472 |
| $\theta_2$ | -0.135088 | 0.135088 |
| $\theta'_1$ | -0.10472 | 0.10472 |
| $\theta'_2$ | -0.135088 | 0.135088 |

Table 2. Range of the states used to set the initial state in the Randomly Varying Initial States condition.

Episodes terminate after 1000 steps or when the angular position of the one of the two poles exceeded the range [-36°, 36°] or the position of the cart exceed the range [-2.4, 2.4] m.

The fitness of the agent corresponds to the fraction of time steps in which the agent maintains the cart and the poles within the allowed position and orientation ranges. Fitness is calculated on the basis of the following equations:

$$f_i = \frac{t}{1000} \quad (10)$$

$$F = \frac{\sum_{i=1}^{8} f_i}{8} \quad (11)$$

where $f_i$ is the fitness of an episode and F is the total fitness.

The evolutionary process is continued until the total number of performed steps exceeds $5*10^7$.

As described above, in this paper only the non-Markovian version of the problem is considered since it represents a relatively difficult task. Results (not shown) on the Markovian double-pole balancing task demonstrate that the task can be easily solved in few evaluations.

As far as the Randomly Varying Initial States condition is concerned, differently from [53], this setup is used to verify the ability of the evolved controllers to deal with the Fixed Initial States condition. In other words, the random initial states are only used to make agents experience a large number of different conditions. This, in turn, should foster their robustness and, therefore, their effectiveness at dealing with the initial states defined in Table 1. The reason behind such design choice depends on the consideration that the performance on the Randomly Varying Initial States condition is strongly biased by chance: some controllers might experience very easy situations to cope with and get high fitness, while others might suffer from hard initial conditions. Evaluating the evolving agents on the Fixed Initial States condition allows the author to discriminate between truly effective and lucky controllers.

The agent and the environment have been simulated by using FARSA [62-63], an open software tool that has been used to successfully transfer results obtained in simulation to hardware for several similar experimental settings [53,64-66].

## 3. Evolutionary algorithms

In this section the evolutionary algorithms used are described in detail.

Before focusing on the algorithms, it is important to underline the different ways of initializing the population depending on the nature of the evolutionary task.

As far as the double-pole balancing task is concerned, the initial population is encoded in a matrix of $\mu$ x $\theta$ integer numbers randomly initialized with a uniform distribution in the range [0, 255] and then converted into values in the range [-8.0,8.0], where $\mu$ corresponds to the number of parents and $\theta$ corresponds to the total number of weights and biases. Offspring are generated by creating a copy of the genotype of the parent and by subjecting each gene to mutation with a MutRate probability. Mutations are made by substituting the gene with a new integer number randomly generated within the range [0, 255] with a uniform distribution.

On the other hand, in the case of the 5-bit parity task the population contains individuals whose genotype is constituted by a vector of integer numbers encoding the function computed by each logic gates and how gates are wired. This approach has been named Cartesian Genetic Programming [59,67]. More specifically, each genotype includes 400 × 3 = 1200 genes specifying the characteristics of the nodes and 1 additional gene representing the identification number of the node used as output for the entire circuit. The inputs are indexed in the range [1–5] and the nodes are indexed in range [6-406]. For each node, one gene bounded in the range [1, 4] indicates the function of the node (1 = OR, 2 = AND, 3 = NAND, 4 = NOR) and two genes bounded in the range $[1, 5 + (L - 1) \times 20]$ indicate the indices of two corresponding inputs of the node. Nodes are arranged in 20 layers receiving inputs only from the previous layers of nodes and from the 5 inputs. L represents the layer of the corresponding logic gate. The value of the last gene, which encodes the node acting as output for the entire circuit, is bounded in the range [6, 406]. Mutations are performed by replacing each integer with a certain probability (MutRate) with a number randomly generated with a uniform distribution in the appropriate range.

### 3.1. Stochastic Steady State

The first evolutionary algorithm is a variant of the standard Steady State evolutionary algorithm [68] called Stochastic Steady State [53].

The Stochastic Steady State (SSS) is a ($\mu$ + $\mu$) evolutionary strategy [69-70] operating on the basis of a population formed by $\mu$ parents. At each generation, each parent generates one offspring, the parent and the offspring are evaluated, and the best $\mu$ individuals are selected as new parents (see Figure 3). It is a method belonging to the class proposed by [71-72]. Differently from previous related methods like the Steady State [68], the authors introduced the possibility of adding noise to the fitness, thus making the selective process stochastic. The noise is a value randomly chosen in the range [-NoiseRange, NoiseRange] with a uniform distribution. When this value is set to 0.0, only the best $\mu$ individuals are allowed to reproduce. The higher the noise, the higher the probability that less fit individuals reproduce. Differently from the original algorithm, the version used in this work (see Figure 3) has been modified by removing the solution refinement during the last $\frac{1}{20}$ period of the evolutionary process (see [53], pp. 10-11).

SSS algorithm:

```
NEvaluations <- 0
    // the genotype of the parents of the first generation is initialized randomly
for p <- 0 to NParents do
    initialize(genome[p])
while (NEvaluations < MaxEvaluations) do
    for p <- 0 to NParents do
        Fitness[p] <- evaluate (p)
        NEvaluations <- NEvaluations + NSteps
        genome[p+NParents] <- genome[p]          // create a copy of parents' genome
        mutate(genome[p+NParents])               // mutate the genotype of the offspring
        Fitness[p+NParents] <- evaluate(genome[p+NParents])
        NEvaluations <- NEvaluations + NSteps
    rank genome[NParents*2] for Fitness[NParents*2]
```

Figure 3. The SSS algorithm.

### 3.2. Hill Climbing

The Hill Climbing (HC) algorithm is a method introduced in [40]. A description is given in Figure 4. As for the SSS algorithm, the population is formed by μ parents. At each generation, each parent generates an offspring (i.e., a mutated copy), and both individuals are evaluated. If the offspring is not worse than its parent, the former replaces the latter in the population. Therefore, differently from the SSS, each individual of the population evolves independently.

HC algorithm:

```
NEvaluations <- 0
   // the genotype of the parents of the first generation is initialized randomly
for p <- 0 to NParents do
   initialize(genome[p])
while (NEvaluations < MaxEvaluations) do
    for p <- 0 to NParents do
          Fitness[p] <- evaluate (p)
          NEvaluations <- NEvaluations + NSteps
          genome[p+NParents] <- genome[p]          // create a copy of parents' genome
          mutate(genome[p+NParents])               // mutate the genotype of the offspring
          Fitness[p+NParents] <- evaluate(genome[p+NParents])
          NEvaluations <- NEvaluations + NSteps
          if Fitness[p+NParents] >= Fitness[p] then
              genome[p] <- genome[p+NParents]
              Fitness[p] <- Fitness[p+NParents]
```

Figure 4. The HC algorithm.

### 3.3. Stochastic Steady State with Hill Climbing

The Stochastic Steady State with Hill Climbing (SSSHC) algorithm is novel algorithm developed by the author. It combines the SSS algorithm with a learning process implemented through a stochastic hill climber [40]. Figure 5 provides the pseudo-code of the SSSHC algorithm. The SSSHC algorithm works as the SSS algorithm until the selection process (Figure 5, lines 1-12), and learning is applied to selected individuals only. During the latter phase, the individuals undergo a refinement process for a fixed number of iterations. At each learning iteration, the currently selected individual is mutated (as it happens during offspring generation, see Figure 5, lines 9 and 18) and the novel individual, referred to as candidate, is evaluated. The currently selected individual is then compared with the candidate. The best performing individual is retained as the currently selected individual. The process is repeated until the given number of learning iterations has been run. Put in other words, variations introduced during learning are inherited [41].

SSSHC algorithm:

```
NEvaluations <- 0
// the genotype of the parents of the first generation is initialized randomly
for p <- 0 to NParents do
    initialize(genome[p])
while (NEvaluations < MaxEvaluations) do
    for p <- 0 to NParents do
        Fitness[p] <- evaluate (p)
        NEvaluations <- NEvaluations + NSteps
        genome[p+NParents] <- genome[p]        // create a copy of parents' genome
        mutate(genome[p+NParents])             // mutate the genotype of the offspring
        Fitness[p+NParents] <- evaluate(genome[p+NParents])
        NEvaluations <- NEvaluations + NSteps
    rank genome[NParents*2] for Fitness[NParents*2]
    for p <- 0 to NParents do
        selectedPop[p] <- genome[p]
        selectedFit[p] <- Fitness[p]
        for iter <- 0 to NLearnIters do
            candidate <- selectedPop[p]        // create a copy of parents' genome
            mutate(candidate)                  // mutate the genotype of the offspring
            candidateFit <- evaluate(candidate)
            NEvaluations <- NEvaluations + NSteps
            if candidateFit > selectedFit[p] then
                selectedPop[p] <- candidate
                selectedFit[p] <- candidateFit
```

Figure 5. The SSSHC algorithm.

## 4. Results

In this section the results obtained in the two different scenarios are presented and discussed. All the statistical analyses have been performed by using Mann-Whitney U test with the application of the Bonferroni correction. A p-value below 0.05 indicates statistical significance. Conversely, when compared conditions are equivalent, the statistical analysis returns a p-value above 0.05.

### 4.1. 5-Bit parity

| No noise | | | Noisy | | |
|---|---|---|---|---|---|
| SSS | HC | SSSHC | SSS | HC | SSSHC |
| 0.974 [0.053] (57655777) | **1.0 [0.0] (8802289)** | **1.0 [0.0] (5432650)** | 0.983 [0.037] (50025341) | **0.999 [0.009] (25986476)** | 0.998 [0.018] (17314381) |

Table 3. Average fitness of the controllers evolved with the SSS, the HC and the SSSHC algorithms. Data obtained by running 50 replications of the experiment. Data in square brackets indicate the standard deviation. Data in circle brackets indicate the average number of evaluations required to find a solution to the problem. Data refer to the best combination of parameters (mutation rate: 1%; population size: 10; noise: 3%).

As indicated by the results reported in Table 3, the SSSHC and HC algorithms outperform the SSS algorithm in the deterministic case (p-value < 0.05). Furthermore, they require a significantly smaller number of evaluations (p-value < 0.05). Noticeably, the SSSHC algorithm is significantly faster than HC (p-value < 0.05). The same result holds with the addiction of noise with respect to both performance (p-value < 0.05) and evaluations (p-value < 0.05). In the noisy condition, the HC algorithm is slightly better than SSSHC, although the difference is not statistically significant (p-value > 0.05). Interestingly, in the deterministic case the SSSHC algorithm manages to solve this relatively complex task with a number of learning iterations in the range [100, 10000] (see Table A5 of the appendix), although it achieves a remarkable performance of 0.99 even with 20-50 learning iterations. The addiction of noise makes the task harder to be solved, requiring a number of learning iterations higher than 200 to get a performance of 0.99 (see table A5 of the appendix).

The better performance of HC over SSS is not surprising. As pointed out in [55], (1+λ)-ES algorithms like HC (where λ=1) display higher performance than (μ + μ)-ES techniques like SSS in this domain. Indeed, the family of parity problems are characterized by high neutrality, i.e. large areas of the search space that can be reached through mutations not affecting the probability to survive and reproduce of an individual [73-76]. Therefore, former methods drive evolution towards such neutral regions of the search space that can ultimately lead to areas with higher fitness, while the latter algorithm tends to explore regions of the search space characterized by high robustness but far from high-fitness areas [55]. The competition between population members observed in (μ + μ)-ES algorithms results in the tendency to perform a local exploration of the search space, thus preventing from discovering optimal solutions to the problem. Moreover, solutions discovered by (1+λ)-ES methods typically contain a higher number of genes playing a functional role than those found by (μ + μ)-ES techniques [55]. The SSSHC exploits the combination of the two different techniques in order to achieve very good performance requiring a considerable small number of evaluation steps (see Table 3).

Figure 6 displays the performance obtained by the different algorithms during evolution. As we can see, SSSHC has a convergence speed noticeably faster than SSS and faster than HC. Concerning the "No noise" case (Figure 6, top), SSSHC manages to find a quasi-optimal solution (i.e., a performance greater or equal to 0.95) after 5225331.2 evaluation steps, HC succeeds in 7323686.4 evaluation steps, while SSS requires 52954150.4 evaluation steps, i.e. more than 10 times the convergence speed of SSSHC and more than 7 times the convergence speed of HC. With respect to the "Noisy" case (Figure 6, bottom), SSSHC discovers a quasi-optimal solution after 17299372.8 evaluation steps, HC takes 20911763.2 evaluation steps, while SSS requires 45649568 evaluation steps, corresponding to more than 2.5 times the convergence speed of SSSHC and more than 2 times the convergence speed of HC.

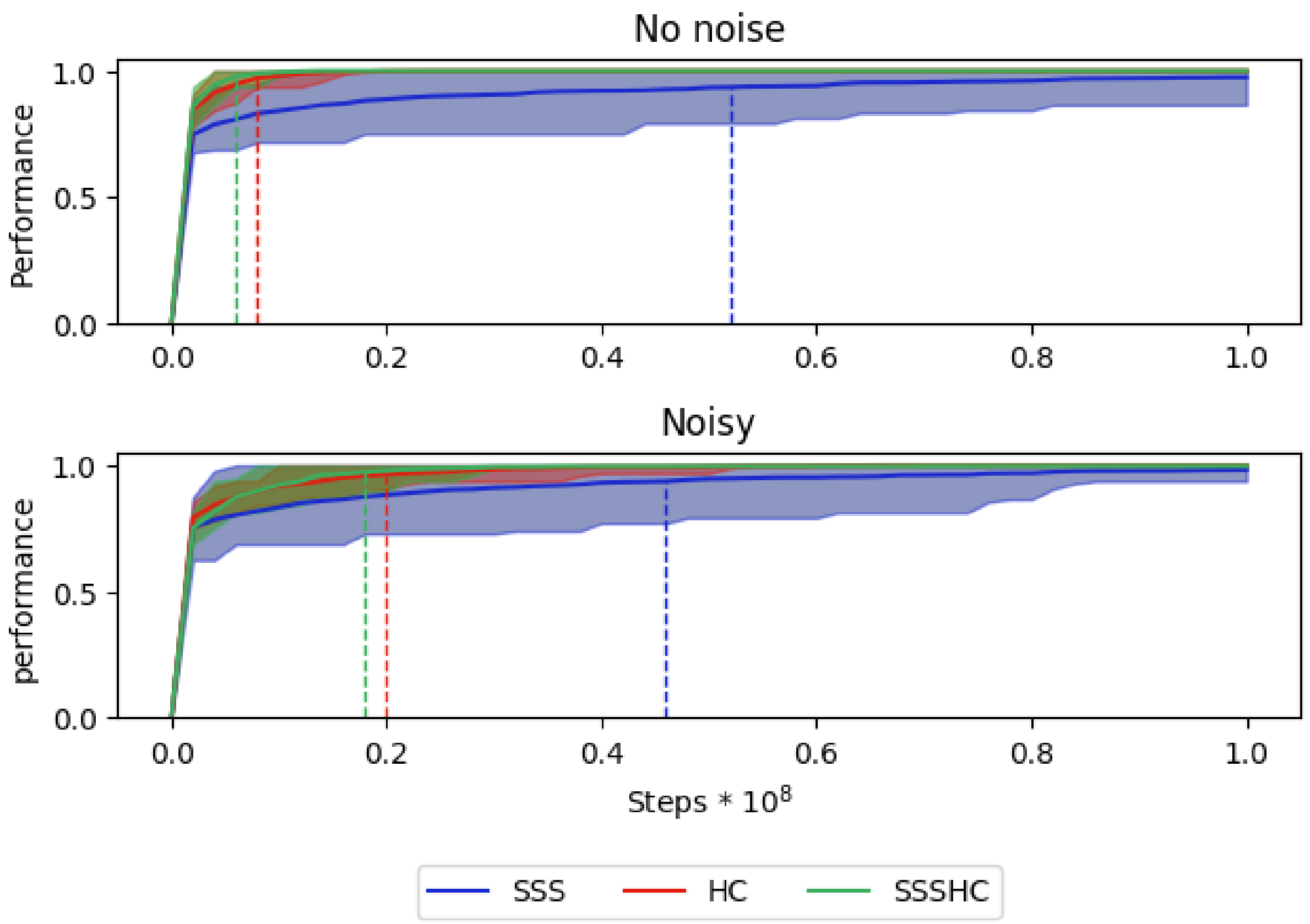

Figure 6. Performance of the SSS, the HC and the SSSHC algorithms with the best combination of parameters (mutation rate: 1%, population size: 10). Top picture refers to experiments without the addiction of noise (learning iterations: 2000). Bottom picture shows results obtained with the addiction of noise (noise: 3%; learning iterations: 5000). Mean and 85% bootstrapped confidence intervals of the mean (shadow area). The vertical dashed line marks the number of steps required by the algorithms to achieve a quasi-optimal solution (i.e., a performance score greater or equal to 0.95).

Figure 7 shows the performance of the SSSHC algorithm depending on the number of learning iterations. Data indicate that the SSSHC algorithm is effective at finding a solution independently of the length of the learning process, at least in this context.

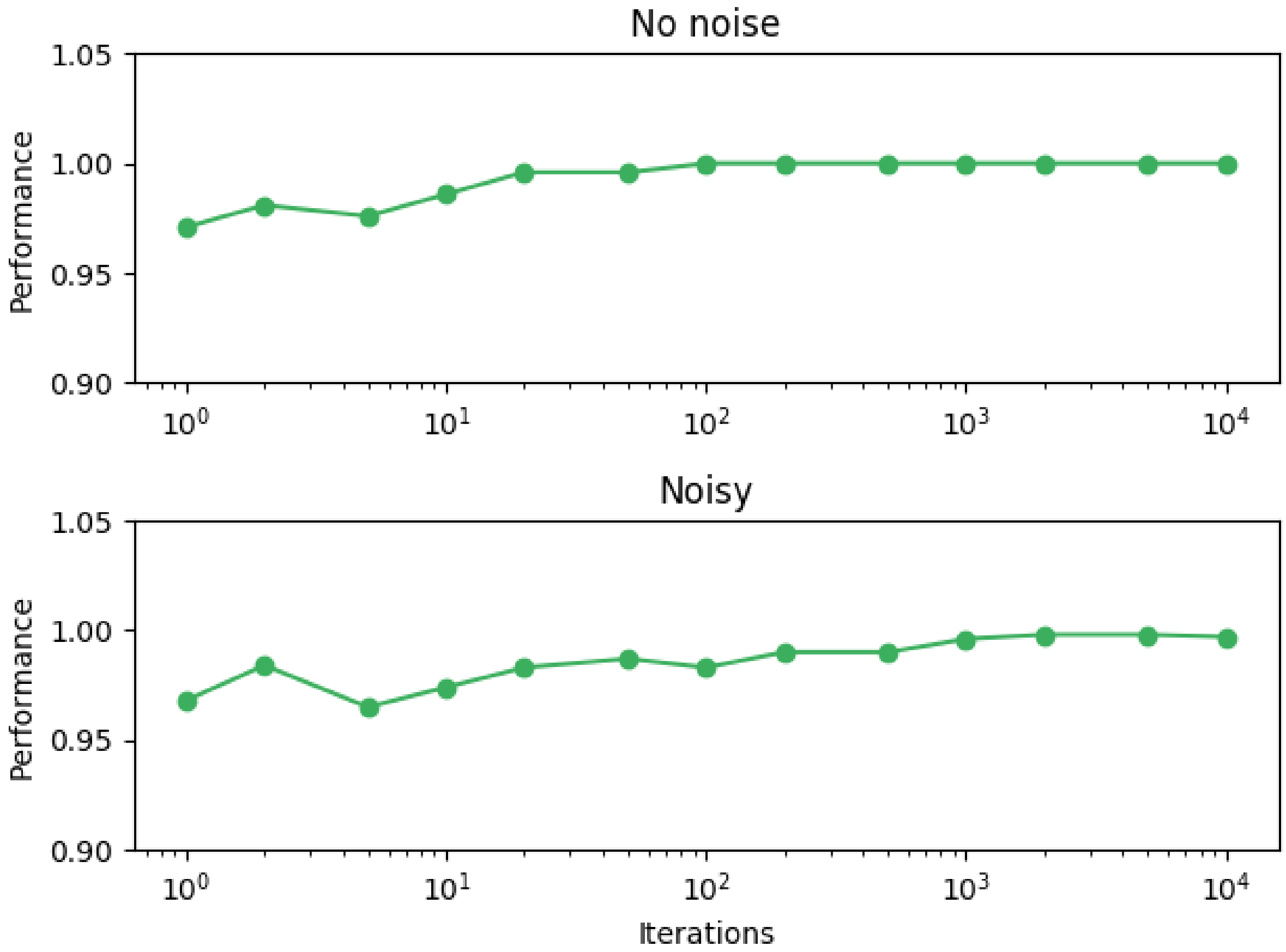

Figure 7. Analysis of the performance of the SSSHC algorithm depending on the different number of learning iterations. Circles indicate the fitness values obtained with the corresponding number of iterations. Top curve shows results without the application of noise, while bottom one displays the performance with the addiction of noise (noise: 3%). Data in the x axis are shown by using a logarithmic scale.

### 4.2. Double-pole balancing

As discussed in section 2.2, the Markovian version of the double-pole balancing problem has not been considered, since it is quite easy to solve. Results obtained in this domain (data not shown) indicate that SSS, HC and SSSHC algorithms successfully manage to solve the problem in both the deterministic case (“No noise”) and the stochastic case (“Noisy”).

#### 4.2.1. Fixed Initial States condition

| No noise | | | Noisy | | |
|---|---|---|---|---|---|
| SSS | HC | SSSHC | SSS | HC | SSSHC |
| **0.996 [0.015] (28095663)** | 0.891 [0.081] (47951820) | 0.991 [0.028] (24576450) | 0.840 [0.100] (47016118) | 0.584 [0.143] (50000000) | **0.950 [0.052] (43856413)** |

Table 4. Average fitness of the controllers evolved with the SSS, the HC and the SSSHC algorithms. Data obtained by running 30 replications of the experiment. Data in square brackets indicate the standard deviation. Data in circle brackets indicate the average number of evaluations required to find a solution to the problem. Data refer to the best combination of parameters (mutation rate: 5%; population size: 200; noise 6%).

As indicated by the results reported in Table 4, the SSS and SSSHC algorithms outperform the HC algorithm (p-value < 0.05) in both conditions. Moreover, the former methods are noticeably faster than the latter (p-value < 0.05) in the "No noise" condition. SSS algorithm slightly outperforms SSSHC algorithm in the "No noise" condition, although the difference is not statistically significant (p-value > 0.05). With respect to the convergence speed, the SSSHC algorithm is faster than the SSS algorithm but the difference is not significant (p-value > 0.05). Considering the addition of noise ("Noisy" condition), the SSSHC remarkably outperforms the SSS algorithm (p-value < 0.05).

Differently from the 5-bit parity task, in this case the HC is significantly worse than the SSS. This implies that strategies in which competition among population members does not play a role are not effective at finding a solution to the problem, at least in this domain. The possibility to compete against other individuals prevents them from getting stuck in sub-optimal solutions. The outcome is even more evident in the "Noisy condition". Indeed, the addition of noise has a disruptive effect on the HC algorithm since the possibility of retaining maladaptive traits increases. Since most of the mutations typically cause a drop in performance and given that evolving individuals do not compete for survival, the combination of these two factors leads to the impossibility to access areas of the search space corresponding to higher fitness. Conversely, the SSS method is less sensitive to the issue, due to the competition among population members triggered by the selection process. The SSSHC benefits from the latter property to avoid being trapped in local minima, while preserving the capability to explore and, possibly, improve the quality of discovered solutions.

The obtained results confirm the hypothesis about the positive influence of noise on learning. Indeed, making the fitness stochastic allows learning to explore more the search space. The retention of maladaptive mutations gives learning the possibility to access areas of the search space that cannot be reached by a deterministic process.

Results in the non-Markovian version of the double-pole suggest a possible relationship between the learning performance and the population size. Differently from the 5-bit parity and the Markovian version of the task (results not shown), in this case the optimal population size is 200, while in the former tasks the optimal value is 10. The hypothesis is that using a large population size does not allow to effectively explore the search space and find optimal solutions. In order to verify it, a control experiment has been run, where only the population size is varied and all other parameters are kept fixed. As a reference task, the non-Markovian double-pole with Fixed States initial condition has been used. The mutation rate has been set to 5%. The number of learning iterations has been set to 5. No noise is added to the learning process. The population size has been varied by using the following values: 10, 20, 50, 100, 200 and 500. Overall, 30 replications of the experiment have been performed, thus evaluating 26400 individuals. The analysis revealed a negative correlation (Spearman correlation test, rho = -0.7808160754210788 significant at $p < 0.01$) between the fitness of the individuals and the population size. In other words, the larger the population size, the lower the performance of the individuals being part of it. This result confirms the hypothesis about the negative effect of large population sizes on the learning process.

### 4.2.2. Randomly Varying Initial States condition

| No noise | | |
|---|---|---|
| SSS | HC | SSSHC |
| 0.903 [0.076] | 0.614 [0.159] | **0.939 [0.066]** |

Table 5. Average fitness of the controllers evolved with the SSS, the HC and the SSSHC algorithms. Data indicate the performance obtained in the Fixed Initial States condition. Data obtained by running 30 replications of the experiment. Data in square brackets indicate the standard deviation. Data refer to the best combination of parameters (mutation rate: 5%; population size: 50).

With regard to the Randomly Varying Initial States condition, as already explained in section 2.2, the ability of the evolved controllers to perform well in the Fixed Initial States condition has been analyzed. In this respect, results reported in Table 5 and Figure 8 indicate that the SSS and SSSHC algorithm significantly outperform the HC algorithm (p-value < 0.05). Furthermore, the SSSHC is better than SSS, although the difference is not statistically significant (p-value > 0.05). The outcomes are in line with those found in section 4.2.1. Therefore, the learning process provides a remarkable effect on the search process, driving it towards higher fitness areas of the search space. It is worth noting that the capability to “generalize” (i.e., to display good performance in the Fixed Initial States condition) is achieved in spite of the intrinsic variability of the task. Further analyses should demonstrate whether or not this property can be extended to other domains.

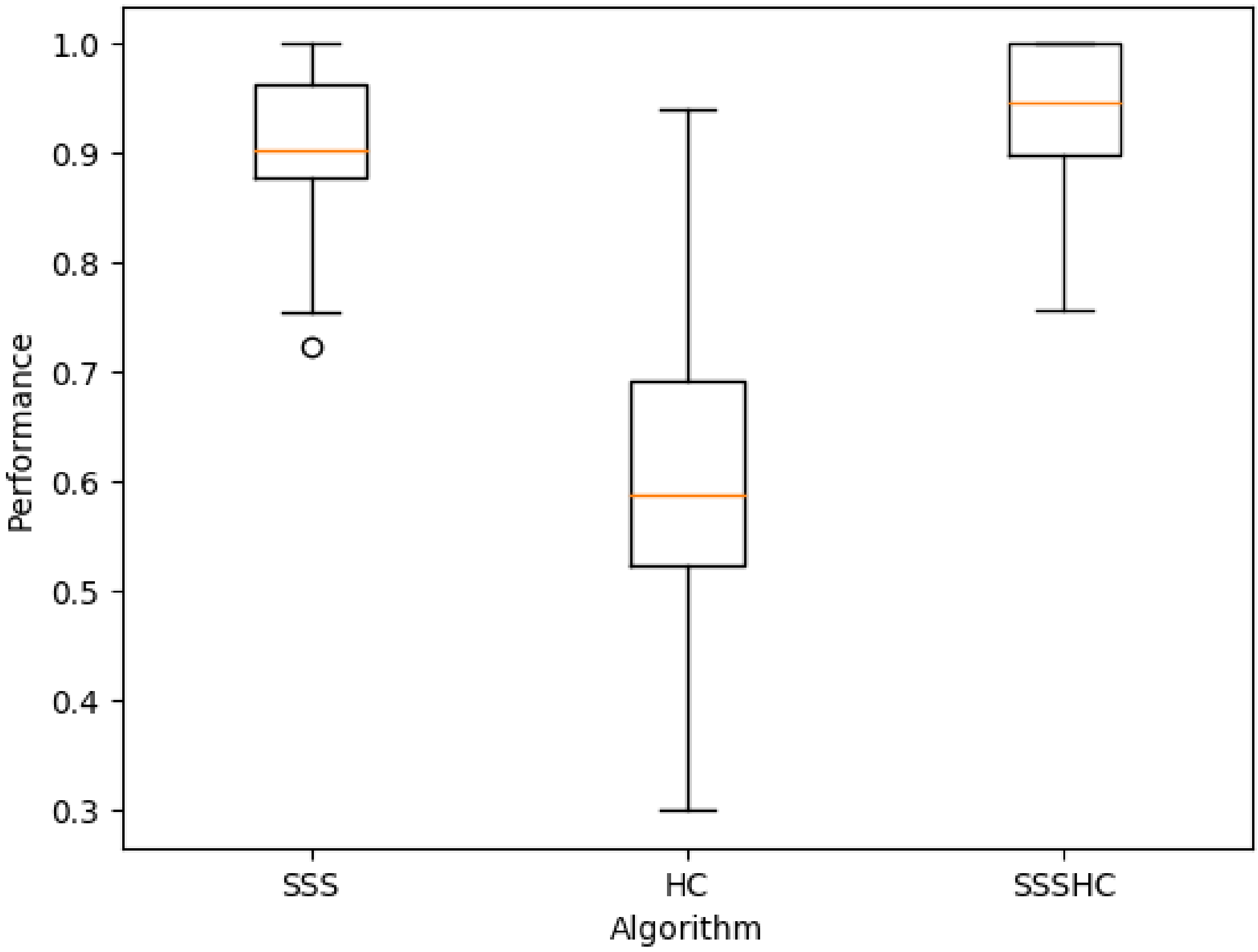

Figure 8. Performance of the SSS, the HC and the SSSHC algorithms in the Randomly Varying Initial States condition. Box represents the inter-quartile range of the data and the horizontal line inside the box marks the median value. The whiskers extend to the most extreme data points within 1.5 times the inter-quartile range from the box. Data obtained by replicating the experiment 30 times with the best combination of parameters (mutation rate: 5%; population size: 50).

## 5. Conclusions

In this paper the benefits and drawbacks of combining evolution and learning, a well-known topic in the research community, have been investigated. Prior works in this area have led to contradictory results, without providing any clue about the actual effect of learning of evolution. Differently from previous approaches, results have been collected in two benchmark tasks, aiming at investigating this interplay on less abstract domains. The hypothesis is that learning provides advantages to evolution, especially when noise is added to the process. A novel algorithm combining learning and evolution, called SSSHC, has been proposed and tested on the well-known 5-bit parity and double-pole balancing tasks.

Results in the presented domains indicate that the combination of evolution and learning is beneficial over the application of evolution alone. Moreover, the advantage is higher when noise is added to the learning and the selection processes. Indeed, the possibility to retain maladaptive traits in order to explore more the search space allows the discovery of areas of higher fitness that cannot be reached through a standard evolutionary process.

It is worth underlining that the results have been obtained by using an unconventional (Lamarckian) learning process retaining adaptive variations in the evolving individual(s). Future works should demonstrate whether the same results hold with a learning process operating only during an individual's lifetime, without affecting the genotype. Furthermore, the domains considered in this work involve limited, or even absent, agent/environment interactions. Future research should clarify whether the combination of learning and evolution achieves better results than evolution alone in conditions where the agent must interact and eventually modify the environment it is situated in.

**Appendix**

**5-bit parity**

**SSS**

| mut_rate/pop_size | 10 | 20 | 50 | 100 |
|---|---|---|---|---|
| 0.01 | **0.974 [0.053]** | 0.939 [0.077] | 0.915 [0.075] | 0.88 [0.106] |
| 0.02 | 0.973 [0.059] | 0.933 [0.077] | 0.922 [0.071] | 0.906 [0.073] |
| 0.05 | 0.955 [0.052] | 0.917 [0.077] | 0.896 [0.081] | 0.898 [0.086] |
| 0.1 | 0.856 [0.083] | 0.832 [0.076] | 0.843 [0.09] | 0.824 [0.074] |
| 0.2 | 0.747 [0.059] | 0.729 [0.047] | 0.723 [0.039] | 0.716 [0.038] |

Table A1. Average fitness of the controllers evolved with the SSS algorithm. Data obtained by running 50 replications of the experiment. Data in brackets indicate the standard deviation

| mut_rate/pop_size | 10 | 20 | 50 | 100 |
|---|---|---|---|---|
| 0.01 | 57655776.64 [35471539.657] | 78376493.44 [32222893.074] | 84238394.88 [29901860.877] | 81910561.92 [31759171.799] |
| 0.02 | 54602154.24 [32553004.011] | 74631771.52 [32766083.903] | 79728540.16 [33550253.227] | 85843382.4 [28171566.198] |
| 0.05 | 75715087.36 [31051535.566] | 83336709.76 [29693240.526] | 88306451.2 [26904713.002] | 86696170.88 [27433135.839] |
| 0.1 | 94481901.44 [18132977.56] | 98077914.88 [9448982.894] | 94318254.08 [16856645.306] | 98964171.52 [7250799.36] |
| 0.2 | 100000000 [0] | 100000000 [0] | 100000000 [0] | 100000000 [0] |

Table A2. Average number of evaluations required to find a solution. Data refer to the controllers evolved with the SSS algorithm. Data obtained by running 50 replications of the experiment. Data in brackets indicate the standard deviation

| | Noise0 | Noise0.01 | Noise0.02 | Noise0.03 | Noise0.04 | Noise0.05 | Noise0.06 | Noise0.07 | Noise0.08 | Noise0.09 | Noise0.1 | Noise0.15 | Noise0.2 |
|---|---|---|---|---|---|---|---|---|---|---|---|---|---|
| mut_rate0.01/ pop_size10 | 0.974 [0.053] | 0.969 [0.063] | 0.969 [0.062] | **0.983 [0.037]** | 0.976 [0.054] | 0.96 [0.07] | 0.969 [0.061] | 0.971 [0.052] | 0.951 [0.072] | 0.965 [0.063] | 0.964 [0.07] | 0.958 [0.067] | 0.945 [0.069] |

Table A3. Average fitness of the controllers evolved with the SSS algorithm with the best combination of parameters. Different levels of noise have been applied. Data obtained by running 50 replications of the experiment. Data in brackets indicate the standard deviation

| | Noise0 | Noise0.01 | Noise0.02 | Noise0.03 | Noise0.04 | Noise0.05 | Noise0.06 | Noise0.07 | Noise0.08 | Noise0.09 | Noise0.1 | Noise0.15 | Noise0.2 |
|---|---|---|---|---|---|---|---|---|---|---|---|---|---|
| mut_rate0.01/pop_size10 | 57655776.64 [35471539.657] | 52773372.16 [35457812.94] | 54974175.36 [37260237.135] | 50025341.44 [35423624.609] | 53352495.36 [34851023.766] | 56632834.56 [35312308.463] | 56116128 [36465003.073] | 53339384.96 [35986901.7054] | 59800429.44 [36766842.906] | 57845848.32 [37715632.476] | 53375178.88 [36374572.333] | 71667905.92 [33530540.938] | 66472620.16 [35725122.778] |

Table A4. Average number of evaluations required to find a solution. Data refer to the controllers evolved with the SSS algorithm with the best combination of parameters. Different levels of noise have been applied. Data obtained by running 50 replications of the experiment. Data in brackets indicate the standard deviation

**SSSHC**

| mut_rate0.01/pop_size10 | LearnIter1 | LearnIter2 | LearnIter5 | LearnIter10 | LearnIter20 | LearnIter50 | LearnIter100 | LearnIter200 | LearnIter500 | LearnIter1000 | LearnIter2000 | LearnIter5000 | LearnIter10000 |
|---|---|---|---|---|---|---|---|---|---|---|---|---|---|
| Noise0 | 0.971 [0.059] | 0.981 [0.036] | 0.976 [0.049] | 0.986 [0.044] | 0.996 [0.02] | 0.996 [0.015] | **1.0 [0.0]** | **1.0 [0.0]** | **1.0 [0.0]** | **1.0 [0.0]** | **1.0 [0.0]** | **1.0 [0.0]** | **1.0 [0.0]** |
| Noise0.03 | 0.968 [0.054] | 0.984 [0.034] | 0.965 [0.064] | 0.974 [0.051] | 0.983 [0.035] | 0.987 [0.042] | 0.983 [0.039] | 0.99 [0.034] | 0.99 [0.025] | 0.996 [0.015] | **0.998 [0.014]** | **0.998 [0.018]** | 0.997 [0.013] |

Table A5. Average fitness of the controllers evolved with the E&L algorithm. Experiments have been run with the best combination of parameters found for the SSS algorithm. Experiments have been run both without noise and with the best amount of noise found for the SSS algorithm. Data obtained by running 50 replications of the experiment. Data in brackets indicate the standard deviation

| mut_rate 0.01/pop_size10 | LearnIter 1 | LearnIter 2 | LearnIter 5 | LearnIter 10 | LearnIter 20 | LearnIter 50 | LearnIter 100 | LearnIter 200 | LearnIter 500 | LearnIter 1000 | LearnIter 2000 | LearnIter 5000 | LearnIter 10000 |
|---|---|---|---|---|---|---|---|---|---|---|---|---|---|
| Noise0 | 54553416.32 [35942853.329] | 50654069.76 [36999291.765] | 48978307.84 [37319084.333] | 39416084.4833385319.28 [34791403.022] | 29982851.2 [33385319.284] | 18887992.32 [24176758.206] | 13652190.08 [20069174.59] | 9908097.28 [15537143.671] | 5736347.52 [5150509.161] | 5799641.6 [3957530.09] | 5432649.6 [2686987.75] | 5546384.64 [2447459.995] | 6331221.12 [2992764.645] |
| Noise0.03 | 53443141.12 [35935029.797] | 53067795.84 [33241888.941] | 51383412.48 [34951684.578] | 52892653.44 [37345237.512] | 47378523.52 [36194083.418] | 39631794.56 [32404020.031] | 37277632 [34827713.428] | 31319713.28 [32354756.987] | 27032099.2 [31356930.517] | 21053765.76 [28566349.967] | 21971118.08 [24984601.063] | 17314380.8 [16447740.793] | 19061704.32 [17165045.329] |

Table A6. Average number of evaluations required to find a solution. Data refer to the controllers evolved with the E&L algorithm. Experiments have been run with the best combination of parameters found for the SSS algorithm. Experiments have been run both without noise and with the best amount of noise found for the SSS algorithm. Data obtained by running 50 replications of the experiment. Data in brackets indicate the standard deviation

**Double-pole balancing**

**Fixed Initial States condition**

| mut_rate/pop_size | 10 | 20 | 50 | 100 | 200 | 500 |
|---|---|---|---|---|---|---|
| 0.01 | 0.088 [0.069] | 0.187 [0.228] | 0.42 [0.356] | 0.515 [0.336] | 0.699 [0.295] | 0.799 [0.159] |
| 0.02 | 0.261 [0.287] | 0.4 [0.345] | 0.664 [0.329] | 0.856 [0.25] | 0.926 [0.121] | 0.898 [0.14] |
| 0.05 | 0.762 [0.294] | 0.925 [0.16] | 0.989 [0.034] | 0.982 [0.061] | **0.996 [0.015]** | 0.979 [0.034] |
| 0.1 | 0.992 [0.027] | 0.988 [0.033] | 0.993 [0.025] | 0.994 [0.017] | 0.99 [0.03] | 0.978 [0.033] |
| 0.2 | 0.958 [0.053] | 0.977 [0.038] | 0.972 [0.045] | 0.967 [0.05] | 0.906 [0.066] | 0.839 [0.073] |

Table A7. Average fitness of the controllers evolved with the SSS algorithm. Data obtained by running 30 replications of the experiment. Data in brackets indicate the standard deviation

| mut_rate/pop_size | 10 | 20 | 50 | 100 | 200 | 500 |
|---|---|---|---|---|---|---|
| 0.01 | 50001719.767 [1654.094] | 50008871.1 [15484.448] | 43237776.467 [14206133.665] | 48376430.333 [6941974.266] | 44935500.767 [10179907.41] | 48879374.6 [6801024.888] |
| 0.02 | 50003683.4 [6567.732] | 46039868.767 [11926746.299] | 46254134.733 [8743061.297] | 34969245.067 [15710975.87] | 42709662.933 [10155102.134] | 45575889.8 [8187062.102] |
| 0.05 | 34835643 [19712121.736] | 30918830.6 [18898235.438] | 21359514 [15370873.958] | 26360703.567 [15358874.296] | 28095663.033 [13273796.979] | 36411198.133 [13622334.511] |
| 0.1 | 16985476 [14810724.587] | 20207577.133 [16127394.493] | 19313581.1 [15788837.491] | 23839773.133 [13309451.3] | 29131395.133 [12191696.413] | 40885220.467 [10959021.963] |
| 0.2 | 39269843.867 [14062396.468] | 35460599.133 [13842691.256] | 35819677.8 [14288714.206] | 39601730.533 [12027354.523] | 47908786.867 [6018875.4] | 50028958.533 [17181.423] |

Table A8. Average number of evaluations required to find a solution. Data refer to the controllers evolved with the SSS algorithm. Data obtained by running 30 replications of the experiment. Data in brackets indicate the standard deviation

| | Noise0 | Noise0.01 | Noise0.02 | Noise0.03 | Noise0.04 | Noise0.05 | Noise0.06 | Noise0.07 | Noise0.08 | Noise0.09 | Noise0.1 | Noise0.15 | Noise0.2 |
|---|---|---|---|---|---|---|---|---|---|---|---|---|---|
| mut_rate0.05/ pop_size200 | **0.996 [0.015]** | 0.838 [0.097] | 0.814 [0.097] | 0.831 [0.01] | 0.801 [0.087] | 0.804 [0.112] | 0.84 [0.1] | 0.839 [0.092] | 0.813 [0.1] | 0.83 [0.087] | 0.801 [0.087] | 0.817 [0.098] | 0.828 [0.091] |

Table A9. Average fitness of the controllers evolved with the SSS algorithm with the best combination of parameters. Different levels of noise have been applied. Data obtained by running 30 replications of the experiment. Data in brackets indicate the standard deviation

| | Noise0 | Noise0.01 | Noise0.02 | Noise0.03 | Noise0.04 | Noise0.05 | Noise0.06 | Noise0.07 | Noise0.08 | Noise0.09 | Noise0.1 | Noise0.15 | Noise0.2 |
|---|---|---|---|---|---|---|---|---|---|---|---|---|---|
| mut_rate0.05/pop_size200 | 28095663.033 [13273796.979] | 49627524.9 [3470701.85] | 50077300.567 [2766451.448] | 48604007.167 [6013811.81] | 49831456.867 [4478650.992] | 50301302.133 [949097.761] | 47016117.733 [10578078.519] | 50271257.367 [1909431.047] | 49435636.833 [5631053.272] | 50024071.2 [2082228.609] | 49905460.467 [3933426.291] | 49227517.9 [5657991.984] | 48989126.867 [5296411.92] |

Table A10. Average number of evaluations required to find a solution. Data refer to the controllers evolved with the SSS algorithm with the best combination of parameters. Different levels of noise have been applied. Data obtained by running 30 replications of the experiment. Data in brackets indicate the standard deviation

**SSSHC**

| mut_rate0.05/pop_size200 | LearnIter1 | LearnIter2 | LearnIter5 | LearnIter10 | LearnIter20 | LearnIter50 | LearnIter100 |
|---|---|---|---|---|---|---|---|
| Noise0 | 0.991 [0.028] | 0.985 [0.028] | 0.974 [0.037] | 0.969 [0.039] | 0.963 [0.049] | 0.948 [0.061] | 0.953 [0.047] |
| Noise0.06 | 0.885 [0.067] | 0.901 [0.075] | 0.936 [0.046] | 0.931 [0.076] | 0.939 [0.056] | 0.950 [0.052] | 0.947 [0.048] |

Table A11. Average fitness of the controllers evolved with the E&L algorithm. Experiments have been run with the best combination of parameters found for the SSS algorithm. Experiments have been run both without noise and with the best amount of noise found for the SSS algorithm. Data obtained by running 30 replications of the experiment. Data in brackets indicate the standard deviation

| mut_rate0.05/pop_size200 | LearnIter 1 | LearnIter 2 | LearnIter 5 | LearnIter 10 | LearnIter 20 | LearnIter 50 | LearnIter 100 |
|---|---|---|---|---|---|---|---|
| Noise0 | 24576449.9 [19896207.317] | 20805706.233 [22703584.519] | 20320573.8 [24875876.509] | 25723404.333 [25714923.851] | 24089354.5 [25764208.284] | 26583520.5 [26588947.701] | 46409139.433 [13411802.525] |
| Noise0.06 | 49479530.167 [5173466.874] | 48477274.833 [7656460.333] | 46758397.067 [9989171.9] | 48841820.7 [5667616.593] | 46716011.967 [9516339.901] | 43856413.3 [14303543.012] | 46774257.2 [12271657.755] |

Table A12. Average number of evaluations required to find a solution. Data refer to the controllers evolved with the E&L algorithm. Experiments have been run with the best combination of parameters found for the SSS algorithm. Experiments have been run both without noise and with the best amount of noise found for the SSS algorithm. Data obtained by running 30 replications of the experiment. Data in brackets indicate the standard deviation

**Random Initial States condition**

**SSS**

| mut_rate/pop_size | 10 | 20 | 50 | 100 | 200 | 500 |
|---|---|---|---|---|---|---|
| 0.01 | 0.779 [0.227] | 0.711 [0.292] | 0.793 [0.245] | 0.845 [0.224] | 0.806 [0.238] | 0.872 [0.127] |
| 0.02 | 0.854 [0.212] | 0.881 [0.181] | 0.883 [0.184] | 0.907 [0.165] | 0.955 [0.075] | 0.907 [0.103] |
| 0.05 | 0.968 [0.073] | 0.956 [0.092] | **0.991 [0.024]** | 0.976 [0.048] | 0.99 [0.024] | 0.955 [0.071] |
| 0.1 | 0.958 [0.11] | 0.95 [0.098] | 0.974 [0.055] | 0.979 [0.048] | 0.954 [0.072] | 0.873 [0.115] |
| 0.2 | 0.763 [0.231] | 0.795 [0.216] | 0.721 [0.2] | 0.615 [0.196] | 0.498 [0.191] | 0.286 [0.132] |

Table A13. Average fitness of the controllers evolved with the SSS algorithm. Data obtained by running 30 replications of the experiment. Data in brackets indicate the standard deviation

| mut_rate/pop_size | 10 | 20 | 50 | 100 | 200 | 500 |
|---|---|---|---|---|---|---|
| 0.01 | 49790696.367 [867449.873] | 48625658.333 [7525890.288] | 50149713.1 [11763148763270.66.899] | 50346989.333 [231743.749] | 50614406.533 [498912.575] | 51044840.133 [795582.668] |
| 0.02 | 46348523.867 [9411799.036] | 48763270.667 [4164967.951] | 46706491.3 [8725737.812] | 47953717.767 [8351406.436] | 49992621.867 [3228328.919] | 51224281.2 [724929.036] |
| 0.05 | 44417856.433 [11933822.024] | 43551810.133 [11284056.44] | 47091485.9 [7844071.626] | 49012520.333 [4734001.551] | 50527518 [308592.49] | 50971748.333 [643773.617] |
| 0.1 | 48330187.467 [5667960.773] | 49384654.267 [3493900.22] | 49934532.433 [699128.755] | 49298191.1 [4384789.493] | 49336571.233 [3593587.801] | 50337967.9 [251965.815] |
| 0.2 | 50010978.033 [8870.574] | 50017604.467 [13045.946] | 50026441.4 [16578.307] | 50049130.133 [24764.599] | 50058213.733 [32488.531] | 50117301.433 [67013.999] |

Table A14. Average number of evaluations required to find a solution. Data refer to the controllers evolved with the SSS algorithm. Data obtained by running 30 replications of the experiment. Data in brackets indicate the standard deviation

| mut_rate/pop_size | 10 | 20 | 50 | 100 | 200 | 500 |
|---|---|---|---|---|---|---|
| 0.01 | 0.648 [0.212] | 0.568 [0.259] | 0.636 [0.235] | 0.631 [0.205] | 0.577 [0.198] | 0.625 [0.133] |
| 0.02 | 0.760 [0.217] | 0.777 [0.181] | 0.777 [0.209] | 0.766 [0.194] | 0.74 [0.141] | 0.685 [0.138] |
| 0.05 | 0.862 [0.140] | 0.878 [0.156] | **0.903 [0.076]** | 0.838 [0.117] | 0.834 [0.076] | 0.772 [0.098] |
| 0.1 | 0.825 [0.145] | 0.787 [0.131] | 0.812 [0.098] | 0.81 [0.104] | 0.775 [0.104] | 0.672 [0.12] |
| 0.2 | 0.484 [0.209] | 0.539 [0.171] | 0.488 [0.16] | 0.391 [0.176] | 0.317 [0.162] | 0.171 [0.08] |

Table A15. Average fitness of the controllers evolved with the SSS algorithm on the fixed initial states. Data obtained by running 30 replications of the experiment. Data in brackets indicate the standard deviation

**SSSHC**

| mut_rate0.05/pop_size50 | LearnIter1 | LearnIter2 | LearnIter5 | LearnIter10 | LearnIter20 | LearnIter50 | LearnIter100 | LearnIter100 |
|---|---|---|---|---|---|---|---|---|
| Noise0 | 0.987 [0.056] | 0.992 [0.034] | 0.996 [0.02] | 1.0 [0.0] | 1.0 [0.0] | 1.0 [0.0] | 1.0 [0.0] | 0.932 [0.063] |

Table A16. Average fitness of the controllers evolved with the E&L algorithm. Experiments have been run with the best combination of parameters found for the SSS algorithm. Experiments have been run both without noise and with the best amount of noise found for the SSS algorithm. Data obtained by running 30 replications of the experiment. Data in brackets indicate the standard deviation

| mut_rate0.05/pop_size50 | LearnIter1 | LearnIter2 | LearnIter5 | LearnIter10 | LearnIter20 | LearnIter50 | LearnIter100 | LearnIter100 |
|---|---|---|---|---|---|---|---|---|
| Noise0 | 45717185.6 [9777947.553] | 44695586.933 [10837929.607] | 44179471.033 [11401559.275] | 49386748.6 [5257182.062] | 47695104.733 [8887242.012] | 46550509.167 [10916586.065] | 48938625.6 [7694881.446] | 41628173.433 [23062620.138] |

Table A17. Average number of evaluations required to find a solution. Data refer to the controllers evolved with the E&L algorithm. Experiments have been run with the best combination of parameters found for the SSS algorithm. Experiments have been run both without noise and with the best amount of noise found for the SSS algorithm. Data obtained by running 30 replications of the experiment. Data in brackets indicate the standard deviation

| mut_rate0.05/pop_size50 | LearnIter1 | LearnIter2 | LearnIter5 | LearnIter10 | LearnIter20 | LearnIter50 | LearnIter100 | LearnIter100 |
|---|---|---|---|---|---|---|---|---|
| Noise0 | 0.913 [0.097] | **0.939 [0.066]** | 0.931 [0.076] | 0.898 [0.048] | 0.931 [0.059] | 0.914 [0.065] | 0.901 [0.067] | 0.233 |

Table A18. Success rate. Data obtained by running 30 replications of the experiment with the best combination of parameters found for the SSS algorithm. Experiments have been run both without noise and with the best amount of noise found for the SSS algorithm.